\documentclass{bmvc2k}

\title{Learning Synergistic Attention for Light Field Salient Object Detection}

\addauthor{Yi Zhang}{yi.zhang1@insa-rennes.fr}{1}
\addauthor{Geng Chen*}{geng.chen.cs@gmail.com}{2}
\addauthor{Qian Chen}{poly@mail.ustc.edu.cn}{3}
\addauthor{YuJia Sun}{thograce@163.com}{4}
\addauthor{Yong Xia}{yxia@nwpu.edu.cn}{2}
\addauthor{Olivier Deforges}{odeforge@insa-rennes.fr}{1}
\addauthor{Wassim Hamidouche}{Wassim.Hamidouche@insa-rennes.fr}{1}
\addauthor{Lu Zhang}{lu.ge@insa-rennes.fr}{1}

\addinstitution{
 Univ Rennes, INSA Rennes, CNRS, IETR (UMR 6164), \\
 France
}
\addinstitution{
 National Engineering Laboratory for Integrated Aero-Space-Ground-Ocean Big Data Application Technology,\\
 School of Computer Science and Engineering, \\
 Northwestern Polytechnical University, \\
 China
}
\addinstitution{
 School of Information Science and Technology,\\
 University of Science and Technology of China,\\
 China
}
\addinstitution{
 School of Computer Science, \\
 Inner Mongolia University,\\
 China 
}

\runninghead{YI ZHANG, GENG CHEN ET AL.}{SA-Net}

\usepackage{amssymb}
\usepackage{overpic}
\usepackage{wrapfig}
\usepackage{wrapfig,lipsum,booktabs}
\usepackage{multirow,longtable}

\newcommand{\tabincell}[2]{\begin{tabular}{@{}#1@{}}#2\end{tabular}}
\newcommand{\secref}[1]{$\S$ \ref{#1}}
\usepackage{rotating}

\def\OurModel{\textit{SA-Net}}
\def\OurCoAttention{\textit{SA}}
\def\OurIntegration{\textit{PF}}
\def\NumOfBaselines{28}

\usepackage[toc,page]{appendix}

\begin{document}

\maketitle

\begin{abstract}
In this work, we propose \textbf{S}ynergistic \textbf{A}ttention \textbf{Net}work (\textbf{\OurModel}) to address the light field salient object detection by establishing a synergistic effect between multi-modal features with advanced attention mechanisms.
Our \OurModel~exploits the rich information of focal stacks via 3D convolutional neural networks, decodes the high-level features of multi-modal light field data with two cascaded synergistic attention modules, and predicts the saliency map using an effective feature fusion module in a progressive manner.
Extensive experiments on three widely-used benchmark datasets show that our \OurModel~outperforms \NumOfBaselines~state-of-the-art models, sufficiently demonstrating its effectiveness and superiority. Our code is available at \url{https://github.com/PanoAsh/SA-Net}.
\end{abstract}

\section{Introduction}\label{sec:intro}

Salient object detection (SOD) is a task aiming to segment the objects that grasp most of the human attention.
It plays a key role in learning human visual mechanism and in various computer vision applications, such as instance segmentation and
video object segmentation.
According to the input modalities, the SOD task can be classified into three categories: 2/3/4D SOD \cite{Zhang2019MemoryorientedDFMoLF}

\begin{wrapfigure}{r}{0.5\textwidth}
\vspace{-8mm}
  \begin{center}
    \includegraphics[width=0.49\textwidth]{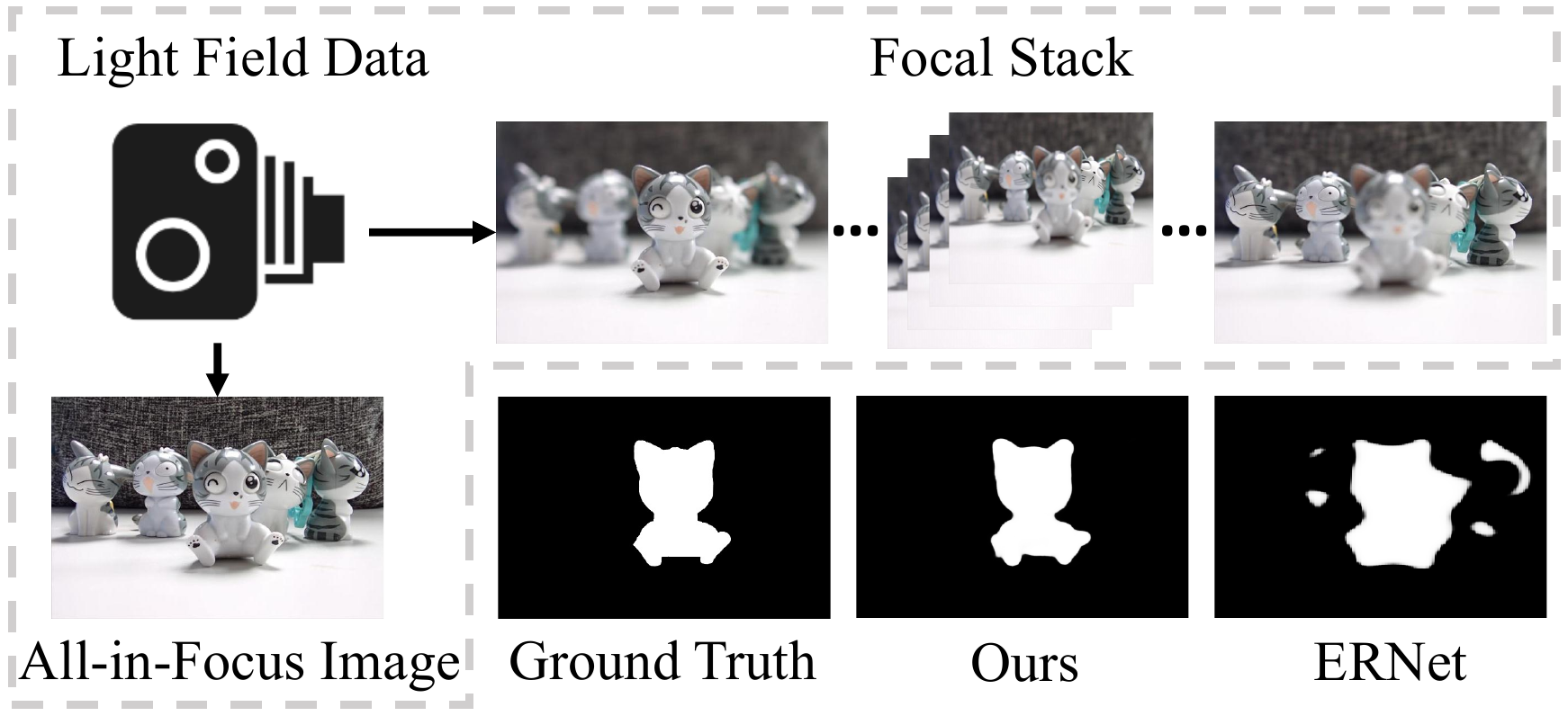}
  \end{center}
	\vspace{-10pt}
  \caption{An example of light field SOD using our \OurModel~and the state-of-the-art model, ERNet \protect\cite{Piao2020ExploitARERNet}.}
 \label{fig:examples}
\end{wrapfigure}

Recently, light field SOD \cite{li2017saliency} (or 4D SOD) has attracted increasing attention owning to the introduction of various light field benchmark datasets, such as DUT-LF \cite{DUTLF}, LFSD \cite{LFSD}, HFUT \cite{HFUT}, DUT-MV \cite{DUTMV}, and Lytro Illum \cite{Lytro}. In addition to all-in-focus (AiF) images, light field datasets \cite{LFSD,HFUT,Lytro} also provide focal stacks (FSs), multi-view images, and depth maps, where the FS is usually known as a series of focal slices focusing at different depths of a given scene while the depth map contains holistic depth information. Unlike 3D (RGB-D) SOD models, which utilize only two modalities, i.e., RGB images and depth maps, the light field SOD models also use multi-view images (e.g., \cite{zhang2020multiMTCNet,DUTMV}), or FSs \cite{DUTLF,Zhang2019MemoryorientedDFMoLF,Zhang2020LFNetLFNet,Piao2020ExploitARERNet} as auxiliary inputs to further improve the performance.
It is worth noting that, most recent FS-based deep learning light field SOD models (e.g., ERNet \cite{Piao2020ExploitARERNet}) have achieved state-of-the-art performance on three widely-used light field benchmark datasets \cite{DUTLF,HFUT,LFSD}.

Despite their advantages, existing works suffer from two major limitations. First, they explore little about the complementarities between AiF images and the FSs.
Existing FS-based methods \cite{DUTLF,Zhang2020LFNetLFNet,Zhang2019MemoryorientedDFMoLF,Piao2020ExploitARERNet} apply only channel attention mechanisms to weight the key feature channels at the decoding stage, to aid the feature fusion between the modalities of AiF and FS. 
%
Considering the fact that salient objects usually appear at specific depths of a given scene, AiF image may include redundant texture details compared to FS, in which a focal slice focuses on a local region at specific depth and blurs the others. New cross-modal fusion strategy, which applies more sophisticated attention mechanisms learning robust cross-modal complementarities, may help solve the issue.
Second, the methods \cite{DUTLF,Zhang2020LFNetLFNet,Zhang2019MemoryorientedDFMoLF,Piao2020ExploitARERNet} all pay little attention to the inter-slice modeling during the encoding stage of FSs. In practice, the AiF images are generated from FSs with a photo-montage technique \cite{agarwala2004interactive}, implying that the former simultaneously depict the spatial details of each local region, while the latter asynchronously focus on different local details along the sequential dimension (Figure \ref{fig:examples}). The relationship between focal slices reflects the context of given scenes as the changes of depth, thus appropriate to be encoded in a progressive manner.
 
To this end, we propose Synergistic Attention Network (\OurModel) to conduct light field SOD with rich information from AiF images and FSs. Specifically, we first employ 3D convolutional neural networks (CNNs) to progressively extract the sequential features from FSs.
At the decoding stage, we propose a synergistic attention (\OurCoAttention) module, where the features from AiF images and FSs are selectively fused and optimized to achieve a synergistic effect for SOD.
Finally, the multi-modal features are fed to our progressive fusion (\OurIntegration) module, which fuses multi-modal features and predicts the saliency map in a progressive manner.
In a nutshell, we provide four main contributions as follows:
\begin{itemize}
    \item We propose the \OurCoAttention~module to decode the high-level features from both AiF images and FSs with a synergistic attention mechanism. Our \OurCoAttention~module exploits the most meaningful information from the multi-modal multi-level features, allowing accurate SOD by taking advantage of light field data.
    \item We introduce a dual-branch backbone to encode the AiF and FS information, simultaneously. To the best of our knowledge, our work is the first attempt to utilize 3D CNNs for the feature extraction of FSs in the field of light field SOD.
    \item We design the \OurIntegration~module to gradually fuse the selective high-level features for the final saliency prediction. 
    \item Extensive experiments demonstrate that our \OurModel~outperforms \NumOfBaselines~state-of-the-art SOD models upon three widely-used light field datasets.
\end{itemize}

\section{Related Works}
In this section, we discuss recent works from the aspects of light field SOD, attention mechanisms, and 3D CNNs.

\subsection{Light Field SOD}\label{sec:related_LFSOD}
\noindent\textbf{Datasets}. To the best of our knowledge, LFSD \cite{LFSD} is the earliest public light field SOD dataset, which provides AiF images, FSs, multi-view (MV) images, depth maps, and micro-lens (ML) images for 100 different scenes. Later, \cite{HFUT} constructed HFUT, which contains 255 AiF images and other light field data types, including FS, MV, ML, and depth. More recent datasets such as DUT-LF \cite{DUTLF} and DUT-MV \cite{DUTMV} contain 1,462 and 1,580 AiF images, respectively. Though being much larger than the early datasets, DUT-LF \cite{DUTLF} does not provide ML and MV images, while DUT-MV \cite{DUTMV} only provides MV images. The most recently proposed dataset, Lytro Illum \cite{Lytro}, provides 640 AiF images as well as the four extra 
forms of light field data. Note that all the five datasets mentioned above are released with SOD ground truth (pixel-wise binary mask) for each AiF image.

\noindent\textbf{Methods}. As the light field SOD is an emerging field, to the best of our knowledge, there are only 18 (11/7 traditional/deep learning-based, respectively) published methods. For traditional ones, the early method \cite{LFSD} conducted light field SOD by considering background and location related prior knowledge. In addition, \cite{Li2015WSC} proposed a unified architecture based on weighted sparse coding. Later methods \cite{Zhang2015SaliencyDILF,Sheng2016RelativeLFRL,Wang2017BIF,li2017saliency,HFUT,Wang2018AccurateSSDDF,Wang2018SalienceGDSGDC} explored and further combined multiple visual cues (e.g., depth, color contrast, light field flows and boundary prior) to detect saliency. Most recent methods \cite{Wang2020RegionbasedDRDFD,Piao2020SaliencyDVDCA} shifted more attention to depth information and employed cellular automata for the saliency detection in light field. With the development of public light field datasets, deep learning-based methods were proposed to conduct SOD task. Specifically, \cite{DUTMV} developed a view synthesis network to detect salient objects by involving MVs. With MVs as inputs, \cite{zhang2020multiMTCNet} further established a unified structure to synchronously conduct salient object and edge detection. Besides, \cite{Lytro} applied DeepLab-v2 for SOD with MLs. As a mainstream, \cite{DUTLF,Zhang2019MemoryorientedDFMoLF,Piao2020ExploitARERNet,Zhang2020LFNetLFNet} all employed ConvLSTM and channel attention mechanisms at the decoding stage to detect salient objects in AiF images and FSs. \cite{DUTLF} and \cite{Zhang2020LFNetLFNet} both modeled the AiF and FS with separate encoder-decoder architectures. Specifically, \cite{DUTLF} added the outputs of the decoders of both the FS branch and AiF branch. \cite{Zhang2020LFNetLFNet} concatenated the outputs of AiF branch and FS branch and used the ConvLSTM \cite{shi2015convolutional} to refine the concatenated features. In \cite{Zhang2019MemoryorientedDFMoLF}, FS and AiF share the same encoder. The extracted multi-modal features were further fused with memory-oriented attention modules. The most recent \cite{Piao2020ExploitARERNet} designed teacher network and student network to encode the FS and AiF respectively, and used ConvLSTM based attention module to facilitate the distillation process between the teacher network and student network.

\subsection{Attention Mechanism}
Generally, there are three main categories of attention mechanisms, including: channel attention \cite{hu2018squeeze}, spatial attention \cite{woo2018cbam}, and self-attention \cite{wang2018non}, which is also a widely known concept in the field of natural language processing. These attention mechanisms can be easily embedded into different CNN-based architectures. Besides, mutual(co)-attention, as a specific type of attention mechanism, has been used in the fields of video object segmentation (e.g., \cite{lu2020zero}), RGB-D SOD (e.g., \cite{LiuS2MA}), etc.
However, mutual attention mechanism has been seldom studied in the field of light field SOD.
As for the SOD task, attention mechanisms are either used for multi-modal fusion (e.g., \cite{LiuS2MA}, \cite{fan2020bbs}) or multi-level feature fusion (e.g., \cite{zhang2018progressive}). Specifically, \cite{zhang2018progressive} proposed a progressive attention guided recurrent network to decode the multi-level features with channel-wise and spatial attention sequentially.

\subsection{3D CNNs}
3D CNNs have proved great competence in modeling spatial-temporal information of video data, thus dominating the video-based detection fields, such as action recognition \cite{carreira2017quo} and video object segmentation \cite{mahadevanmaking}.

Recently, RD3D \cite{chen2021rd3d} was proposed to address the task of 3D (RGB-D) SOD by using a 3D CNN-based encoder-decoder structure, and achieved promising performance on widely-used RGB-D SOD benchmarks.
As for light field SOD, MTCNet \cite{zhang2020multiMTCNet} applied 3D CNN-based encoder to extract the depth features from multi-view images. The rich high-level features gained from 3D CNNs were then used to infer depth maps and facilitate the SOD task, synchronously.
Since FSs are sequences of focal slices focusing at different depths, learning FSs' features via 3D CNNs possesses great potential to boost the model performance for light field SOD, but so far lacks investigation.

\begin{figure*}[t!]
	\centering
	{\includegraphics[trim={0 1.2cm 0 0},clip,width=0.95\textwidth,page=1]{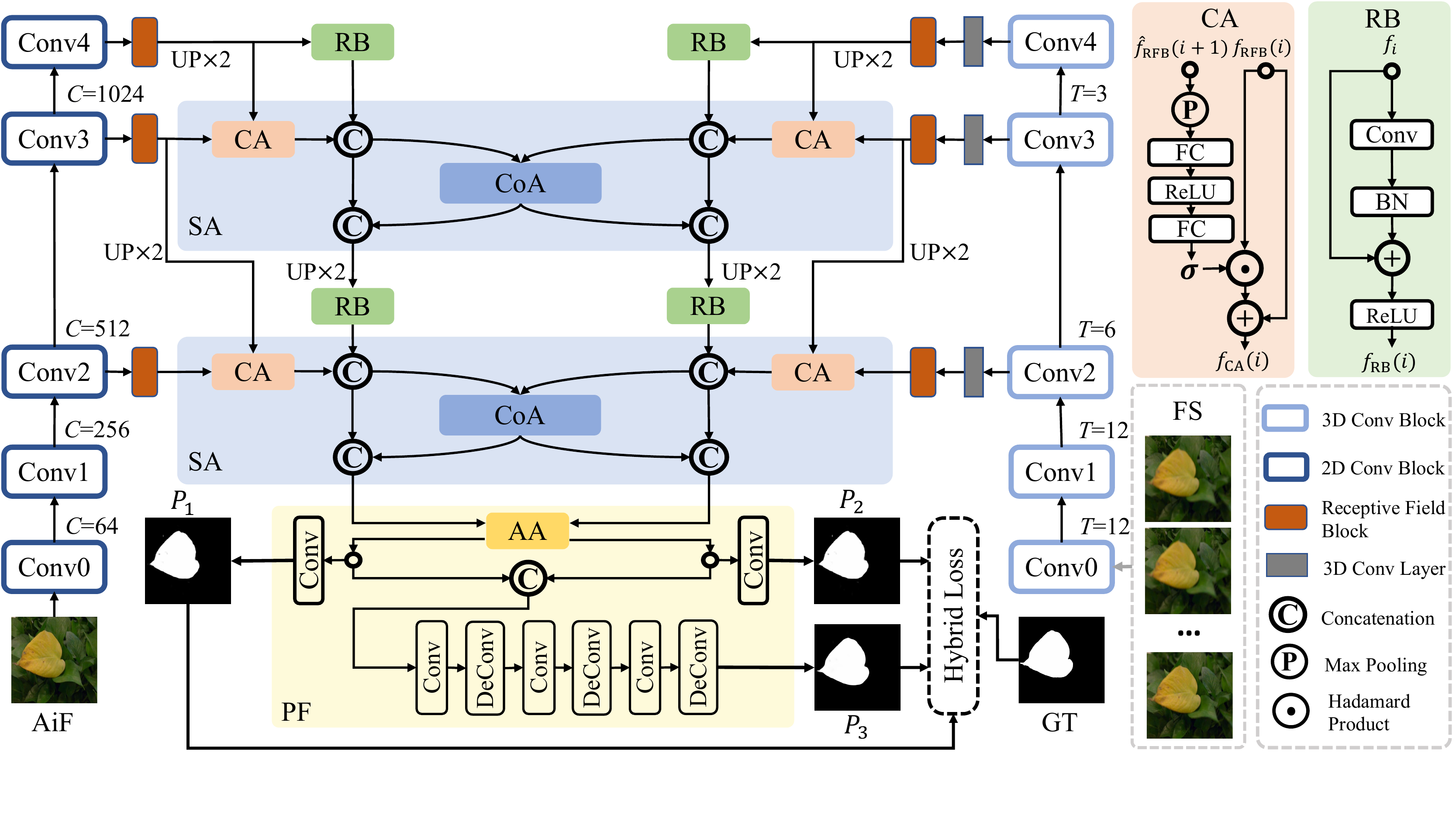}} 
    \vspace{-8pt}
	\caption{An overview of our \textbf{\OurModel}. Multi-modal multi-level features extracted from our multi-modal encoder are fed to two cascaded synergistic attention (\OurCoAttention) modules followed by a progressive fusion (\OurIntegration) module. The short names in the figure are detailed as follows:
	CoA = co-attention component. CA = channel attention component. AA = AiF-induced attention component.
	RB = residual block.
	$\text{P}_{{n}}$ = the $n$th saliency prediction.
	(De)Conv = (de-)convolutional layer.
	BN = batch normalization layer. FC = fully connected layer.}
    \label{fig:model_main}
\end{figure*}

\section{Synergistic Attention Network}
In \OurModel, we exploit rich cross-modal complementary information with channel attention and co-attention mechanisms to achieve a synergistic effect between multi-level AiF and FS features.
In addition, to capture the inter-slice information of FS, we employ 3D CNNs to extract rich features from FSs.
Figure~\ref{fig:model_main} shows an overview of our \OurModel, which consists of three major components, including a multi-modal encoder consisting of 2D and 3D CNNs (\secref{sec:encoder}), two cascaded \OurCoAttention~modules (\secref{sec:SA-module}), and a \OurIntegration~module (\secref{sec:PF-module}).

\subsection{Multi-Modal Encoder}\label{sec:encoder}
As shown in Figure \ref{fig:model_main}, the encoder of our network is a dual-branch architecture for synchronous feeding of the two modalities, i.e., AiF images and FSs.
For the 2D branch, we encode an input AiF image
with a group of convolutional blocks.
On the other hand, FS
is represented as a 4D tensor with the last dimension $T$ denoting the number of focal slices.
We encode the FS with a stack of 3D convolutional blocks, which are able to jointly capture the rich intra- and inter-slice information for accurate SOD.
Note that the same setting ($T = 12$) as in \cite{Piao2020ExploitARERNet} is adopted in our 3D branch, and a zero-padding strategy is applied to the FS with less than 12 focal slices.

\subsection{Synergistic Attention Module}\label{sec:SA-module}
As high-level features tend to reserve the essential cues (e.g., location, shape) of salient objects while the low-level ones contain relative trivial information (e.g., edge) \cite{wu2019cascaded}, our decoder only integrates high-level features to avoid redundant computational complexity. Specifically, we use $\{f_{i}^{\text{2D}}\}_{i=2}^{4}$ and $\{f_{i}^{\text{3D}}\}_{i=2}^{4}$ to denote the high-level AiF and FS features extracted from the 2D and 3D CNNs of our dual-branch backbone network (\secref{sec:encoder}).



\begin{figure*}[t!]
	\centering
    \begin{overpic}[width=0.99\textwidth]{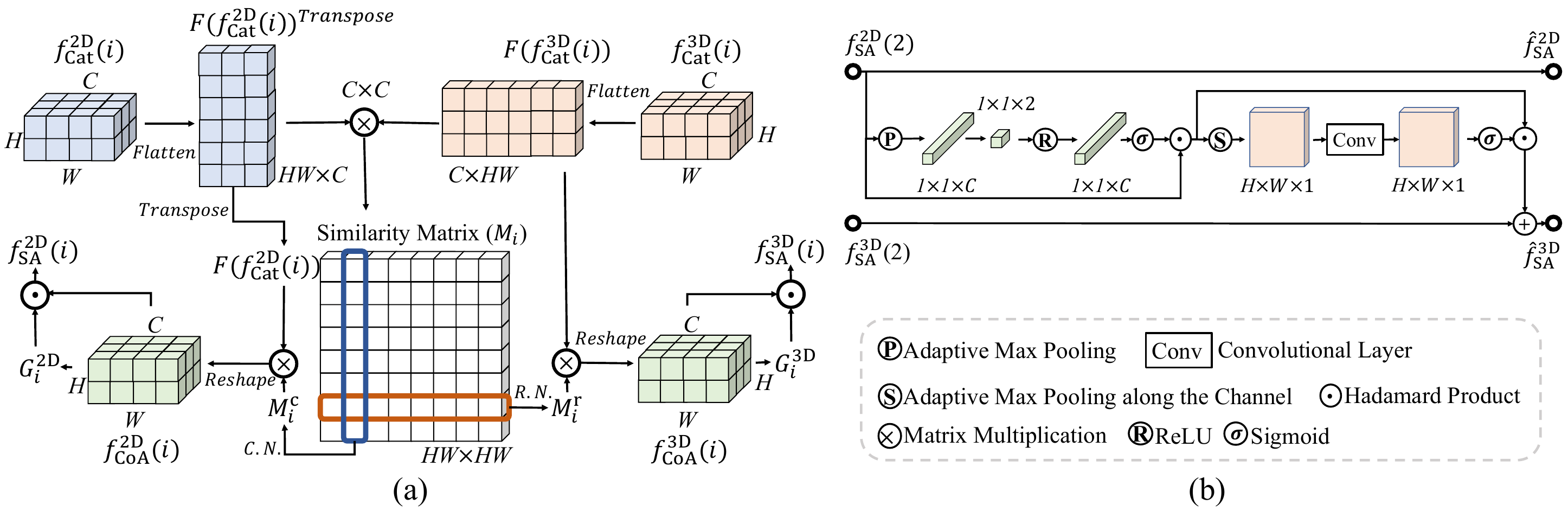}
    \end{overpic}
	\caption{(a) The architecture of the co-attention (CoA) module. C.N. = column-wise normalization. R.N. = row-wise normalization. (b) The architecture of AiF-induced attention component.}
    \label{fig:model_sub}
\end{figure*}

\noindent\textbf{Multi-Level Attention}. As shown in Figure \ref{fig:model_main}, a receptive field block (RFB) \cite{wu2019cascaded} is first employed to enrich the global context information for each convolution block. Taking the AiF branch as an example, the adjacent high-level features from the encoder are then combined with a channel attention (CA) mechanism from \cite{hu2018squeeze}, i.e.,
\begin{equation}\label{equ:MA}
 \begin{split}
   f_{\text{CA}}^{\text{2D}}(i) = \sigma(FC(ReLU(FC(P(\hat{f}_{\text{RFB}}^{\text{2D}}(i+1)))))) \odot f_{\text{RFB}}^{\text{2D}}(i) + f_{\text{RFB}}^{\text{2D}}(i),
   \end{split}
\end{equation}
where $f_{\text{RFB}}^{\text{2D}}(i)$ represents the $i$th level features provided by RFB; $\hat{f}_{\text{RFB}}^{\text{2D}}(i+1)$ is the up-sampled version of $f_{\text{RFB}}^{\text{2D}}(i+1)$; $\sigma(\cdot)$, $FC(\cdot)$, $P(\cdot)$, and $\odot$ denotes the Sigmoid function, fully connected layer, max pooling, and Hadamard product, respectively. The resulting feature $f_{\text{CA}}^{\text{2D}}(i)$ is further concatenated with the upper level feature $f_{\text{RB}}^{\text{2D}}(i+1)$ provided by a residual block (RB) for the feature $f_{\text{Cat}}^{\text{2D}}(i)$, which is one of the pair-wise inputs ($\{f_{\text{Cat}}^{\text{2D}}(i), f_{\text{Cat}}^{\text{3D}}(i)\}$) for the second stage of our \OurCoAttention~module.
Note that the FS branch follows the consistent procedure as in AiF's since two branches are symmetric.

\noindent\textbf{Multi-Modal Attention}. Inspired by a mutual attention mechanism \cite{lu2020zero} used for cross-frame feature fusion in the field of video object segmentation, the high-level feature interaction between the two modalities is conducted with two cascaded co-attention (CoA) modules (Figure \ref{fig:model_main}). As shown in Figure \ref{fig:model_sub} (a), given the pair-wise features $\{f_{\text{Cat}}^{\text{2D}}(i), f_{\text{Cat}}^{\text{3D}}(i)\}$ at $i$th layer as inputs, a similarity matrix $M_i$ can be computed as:
\begin{equation}\label{equ:CoA_1}
   M_i = F(f_{\text{Cat}}^{\text{2D}}(i))^{\text{T}} \otimes F(f_{\text{Cat}}^{\text{3D}}(i)),
\end{equation}
where $F(\cdot)$ represents a flatten operation reshaping the 3D feature matrix
$f_{\text{Cat}}^{\text{2D}}(i) \in \mathbb{R}^{H\times W\times C}$
to a 2D one with a dimension of $HW\times C$,
$\otimes$ denotes matrix multiplication. Note that we do not apply extra weight matrix as in \cite{lu2020zero} to compute $M_i$, since the CoA module aims at fusing the cross-modal features with equally assigned attention.
The $M_i$ is then column-/row-wisely normalized via:
\begin{equation}\label{equ:CoA_2}
\begin{split}
   M_i^{\text{c}} &= Softmax(M_i) \in [0, 1]^{{HW \times HW}},\\
   M_i^{\text{r}} &= Softmax(M_i^{\text{T}}) \in [0, 1]^{{HW \times HW}},
 \end{split}
\end{equation}
where $Softmax(\cdot)$ normalizes each column of the similarity matrix. Therefore, the co-attention-based pair-wise features ($\{f_{\text{CoA}}^{\text{2D}}(i), f_{\text{CoA}}^{\text{3D}}(i)\}$) at $i$th layer are further defined as:
\begin{equation}\label{equ:CoA_3}
\begin{split}
  f_{\text{CoA}}^{\text{2D}}(i) &= R(f_{\text{Cat}}^{\text{2D}}(i) \otimes M_i^{\text{c}}) \in [0, 1]^{{H \times W \times C}},\\
  f_{\text{CoA}}^{\text{3D}}(i) &= R(f_{\text{Cat}}^{\text{3D}}(i) \otimes M_i^{\text{r}}) \in [0, 1]^{{H \times W \times C}},
   \end{split}
\end{equation}
where $R(\cdot)$ reshapes the given matrix from a dimension of $C \times HW$ to $H \times W \times C$. A self-gate mechanism \cite{lu2020zero} is further employed to automatically learn the co-attention confidences ($G_{i}^{\text{2D}}, G_{i}^{\text{3D}}$) for $f_{\text{CoA}}^{\text{2D}}(i)$ and $f_{\text{CoA}}^{\text{3D}}(i)$. Therefore the final outputs $\{f_{\text{SA}}^{\text{2D}}(i), f_{\text{SA}}^{\text{3D}}(i)\}$ of our \OurCoAttention~module at $i$th layer are computed as:
\begin{equation}\label{equ:CoA_Fin}
   f_{\text{SA}}^{\text{2D}}(i) = G^{\text{2D}}_i \odot f_{\text{CoA}}^{\text{2D}}(i)~\text{and}~f_{\text{SA}}^{\text{3D}}(i) = G^{\text{3D}}_i \odot f_{\text{CoA}}^{\text{3D}}(i),
\end{equation}
where the co-attention confidence $G^{\text{2D}}_i = \sigma(Conv(f_{\text{CoA}}^{\text{2D}}(i)))$ with $Conv(\cdot)$ denoting a convolutional layer.

By combining the CA and CoA module, our \OurCoAttention~module is particularly effective in exploiting the multi-level and multi-modal complementary information, which, therefore, provides significantly improved performance, as demonstrated by our ablation studies in \secref{sec:ab}.

\subsection{Progressive Fusion Module}\label{sec:PF-module}
To obtain the final prediction, we further add a \OurIntegration~module to gradually up-sample the selective high-level features provided by our \OurCoAttention~module (Figure \ref{fig:model_main}). Specifically, we first balance the FS and AiF features with an AiF-induced attention (AA) component (Figure \ref{fig:model_sub} (b)) before the final fusion of the two modalities.
The AA component follows the same procedure applied in RGB-D fusion \cite{fan2020bbs}, i.e., unifying the channel and spatial attention by computing:
\begin{equation}\label{equ:AA}
\begin{split}
   \hat{f}_{\text{SA}}^{\text{2D}}=f_{\text{SA}}^{\text{2D}}(2)~\text{and}~\hat{f}_{\text{SA}}^{\text{3D}} = SA(CA(f_{\text{SA}}^{\text{2D}}(2))) + f_{\text{SA}}^{\text{3D}}(2),
   \end{split}
\end{equation}
where $CA(\cdot)$ and $SA(\cdot)$ denote spatial and channel attention components, respectively. 
We then concatenate the balanced cross-modal features and feed them to a deconvolutional block for the final prediction $P_3$, i.e.,
\begin{equation}\label{equ:PF}
    P_3 = DB(Cat(\hat{f}_{\text{SA}}^{\text{2D}}, \hat{f}_{\text{SA}}^{\text{3D}})),
\end{equation}
where $Cat(\cdot)$ denotes the concatenation operation, and $DB(\cdot)$ represents a deconvolutional block consisting of three deconvolutional layers \cite{fan2020bbs} and convolutional layers that are organized in a cascaded manner (Figure \ref{fig:model_main}).

\subsection{Loss Function}
As shown in Figure~\ref{fig:model_main}, our model predicts three saliency maps: $\{P_n\}_{n=1}^{3}\in [0,1]$. Let $G \in \{0,1\}$ denotes the ground-truth saliency map, we jointly optimize the three-way predictions by defining a hybrid loss $\ell$:
\begin{equation}\label{equ:loss}
   \ell = \sum_{n=1}^{N}\ell_{\text{BCE}}(P_{n}, G) + \ell_{\text{IoU}}(P_{n}, G) + \ell_{\text{EM}}(P_{n}, G),
\end{equation}
where $\ell_{\text{BCE}}$ and $\ell_{\text{IoU}}$ denote Binary Cross Entropy (BCE) and Intersection over Union (IoU) loss, respectively; the loss $\ell_{\text{EM}} = 1 - E_\phi$ with $E_\phi$ denoting E-Measure \cite{Fan2018Enhanced}.

\subsection{Implementation Details}
Our \OurModel~is implemented in PyTorch and optimized with Adam algorithm \cite{kingma2014adam}.
The backbone of \OurModel~is based on a 2D standard ResNet50 for AiF images and an inflated 3D ResNet50 \cite{carreira2017quo} for FSs.
The 2D convolution layers in our backbone are initialized with ImageNet-pretrained ResNet50, while the 3D convolutional layers are initialized with a 2D weight transfer strategy \cite{carreira2017quo}.
During the training stage, the batch size is set to 2, the learning rate is initialized as 1e-5 and decreased by 10$\%$ when training loss reaches a flat. It takes about 14 hours to train the proposed model based on a platform consists of Intel$^\circledR$ i9-7900X CPU@3.30GHz and one TITAN XP.

\section{Experiments}

\begin{table*}[t]
	\centering
	\renewcommand{\arraystretch}{1.0}
	\setlength\tabcolsep{6pt}
	\resizebox{0.99\textwidth}{!}{
		\begin{tabular}{ll||cccc||cccc||cccc}
			\hline
			\toprule
			\multirow{2}{*}{{Types}} & \multirow{2}{*}{{Models}} 
			& \multicolumn{4}{c||}{\tabincell{c}{DUT-LF~\cite{DUTLF}}} 
			& \multicolumn{4}{c||}{\tabincell{c}{HFUT~\cite{HFUT}}} 
			& \multicolumn{4}{c}{ \tabincell{c}{LFSD~\cite{LFSD}}} \\
			\cline{3-14}
			& & $F_\beta\uparrow$ & $S_\alpha\uparrow$ & $E_\phi\uparrow$ & $M\downarrow$
			& $F_\beta\uparrow$ & $S_\alpha\uparrow$ & $E_\phi\uparrow$ & $M\downarrow$
			& $F_\beta\uparrow$ & $S_\alpha\uparrow$ & $E_\phi\uparrow$ & $M\downarrow$
			\\
			\hline
			\multirow{8}{*}{{4D}}	
    &	SANet	&	\textbf{0.920}	&	\textbf{0.918}	&	\textbf{0.954}	&	\textbf{0.032}	&	\textbf{0.736}	&	\textbf{0.784}	&	\textbf{0.849}	&	\textbf{0.078}	&	\textbf{0.844}	&	\textbf{0.841}	&	\textbf{0.889}	&	\textbf{0.074}	\\
~	&	ERNetT~\cite{Piao2020ExploitARERNet}	&	{0.889}	&	{0.899}	&	{0.943}	&	{0.040}	&	0.705	&	0.777	&	{0.831}	&	0.082	&	{0.842}	&	{0.838}	&	\textbf{0.889}	&	{0.080}	\\
~	&	ERNetS~\cite{Piao2020ExploitARERNet}	&	0.838	&	0.848	&	0.916	&	0.061	&	0.651	&	0.736	&	0.824	&	0.085	&	0.721	&	0.726	&	0.820	&	0.137	\\
~	&	{DLFS}~\cite{DUTMV}	&	0.801	&	0.841	&	0.891	&	0.076	&	0.615	&	0.741	&	0.783	&	0.098	&	0.715	&	0.737	&	0.806	&	0.147	\\
~	&	LFS$^{\star}$~\cite{li2017saliency}	&	0.484	&	0.563	&	0.728	&	0.240	&	0.430	&	0.579	&	0.686	&	0.205	&	0.740	&	0.680	&	0.771	&	0.208	\\
~	&	MCA$^{\star}$~\cite{HFUT}	&	-	&	-	&	-	&	-	&	-	&	-	&	-	&	-	&	0.815	&	0.749	&	0.841	&	0.150	\\
~	&	WSC$^{\star}$~\cite{Li2015WSC}	&	-	&	-	&	-	&	-	&	-	&	-	&	-	&	-	&	0.706	&	0.706	&	0.794	&	0.156	\\
~	&	DILF$^{\star}$~\cite{Zhang2015SaliencyDILF}	&	0.641	&	0.705	&	0.805	&	0.168	&	0.555	&	0.695	&	0.736	&	0.131	&	0.728	&	0.755	&	0.810	&	0.168	\\
\hline	

\multirow{9}{*}{{3D}} 
    & S2MA~\cite{LiuS2MA} & 0.754 & 0.787 & 0.841 & 0.103 & 0.647 & 0.761 & 0.787 & 0.100 & 0.819 & 0.837 & 0.863 & 0.095  \\
~   & D3Net~\cite{fan2020rethinking} & 0.790 & 0.822 & 0.869 & 0.084 & 0.692 & 0.778 & 0.827 & 0.080 & 0.804 & 0.825 & 0.853 & 0.095 \\
~   &	CPFP~\cite{zhao2019contrast}	&	0.730	&	0.741	&	0.808	&	0.101	&	0.594	&	0.701	&	0.768	&	0.096	&	0.524	&	0.599	&	0.669	&	0.186	\\
~	&	TANet~\cite{Chen2019TANet}	&	0.771	&	0.803	&	0.861	&	0.096	&	0.638	&	0.744	&	0.789	&	0.096	&	0.804	&	0.803	&	0.849	&	0.112	\\
~	&	MMCI~\cite{Chen2019MMCI}	&	0.750	&	0.785	&	0.853	&	0.116	&	0.645	&	0.741	&	0.787	&	0.104	&	0.796	&	0.799	&	0.848	&	0.128	\\
~	&	PDNet~\cite{zhu2019pdnet}	&	0.763	&	0.803	&	0.864	&	0.111	&	0.629	&	0.770	&	0.786	&	0.105	&	0.780	&	0.786	&	0.849	&	0.116	\\
~	&	PCA~\cite{chen2018progressively}	&	0.762	&	0.800	&	0.857	&	0.100	&	0.644	&	0.748	&	0.782	&	0.095	&	0.801	&	0.807	&	0.846	&	0.112	\\
~	&	CTMF~\cite{Han2017CTMF}	&	0.790	&	0.823	&	0.881	&	0.100	&	0.620	&	0.752	&	0.784	&	0.103	&	0.791	&	0.801	&	0.856	&	0.119	\\
~	&	DF~\cite{qu2017rgbd}	&	0.733	&	0.716	&	0.838	&	0.151	&	0.562	&	0.670	&	0.742	&	0.138	&	0.756	&	0.751	&	0.816	&	0.162	\\
\hline	

\multirow{12}{*}{{2D}} 
     & F3Net \cite{wei2020f3net} & 0.882 & 0.888 & 0.900 & 0.057 & 0.718 & 0.777 & 0.815 & 0.095 & 0.797 & 0.806 & 0.824 & 0.106 \\
~   & GCPANet~\cite{GCPANet} & 0.867 & 0.885 & 0.898 &  0.064 & 0.691 & 0.777 & 0.799 & 0.105 & 0.805 & 0.822 & 0.809 & 0.097\\ 
~   &	EGNet~\cite{zhao2019egnet}	&	0.870	&	0.886	&	0.914	&	0.053	&	0.672	&	0.772	&	0.794	&	0.094	&	0.762	&	0.784	&	0.776	&	0.118	\\
~	&	PoolNet~\cite{liu2019simple}	&	0.868	&	0.889	&	0.919	&	0.051	&	0.683	&	0.776	&	0.802	&	0.092	&	0.769	&	0.800	&	0.786	&	0.118	\\
~	&	PAGRN~\cite{zhang2018progressive}	&	0.828	&	0.822	&	0.878	&	0.084	&	0.635	&	0.717	&	0.773	&	0.114	&	0.725	&	0.727	&	0.805	&	0.147	\\
~	&	C2S~\cite{li2018contour}	&	0.791	&	0.844	&	0.874	&	0.084	&	0.650	&	0.763	&	0.786	&	0.111	&	0.749	&	0.806	&	0.820	&	0.113	\\
~	&	R$^{3}$Net~\cite{deng2018r3net}	&	0.783	&	0.819	&	0.833	&	0.113	&	0.625	&	0.727	&	0.728	&	0.151	&	0.781	&	0.789	&	0.838	&	0.128	\\
~	&	Amulet~\cite{zhang2017amulet}	&	0.805	&	0.847	&	0.882	&	0.083	&	0.636	&	0.767	&	0.760	&	0.110	&	0.757	&	0.773	&	0.821	&	0.135	\\
~	&	UCF~\cite{zhang2017learning}	&	0.769	&	0.837	&	0.850	&	0.107	&	0.623	&	0.754	&	0.764	&	0.130	&	0.710	&	0.762	&	0.776	&	0.169	\\
~	&	SRM~\cite{wang2017stagewise}	&	0.832	&	0.848	&	0.899	&	0.072	&	0.672	&	0.762	&	0.801	&	0.096	&	0.827	&	0.826	&	0.863	&	0.099	\\
~	&	NLDF~\cite{luo2017non}	&	0.778	&	0.786	&	0.862	&	0.103	&	0.636	&	0.729	&	0.807	&	0.091	&	0.748	&	0.745	&	0.810	&	0.138	\\
~	&	DSS~\cite{hou2017deeply}	&	0.728	&	0.764	&	0.827	&	0.128	&	0.626	&	0.715	&	0.778	&	0.133	&	0.644	&	0.677	&	0.749	&	0.190	\\

			\bottomrule
		\end{tabular}
	}
    \vspace{+10pt}
	\caption{Quantitative results for different models on three benchmark datasets. The best scores are in \textbf{boldface}. We train and test our \OurModel~with the settings that are consistent with \protect\cite{Piao2020ExploitARERNet}, which is the state-of-the-art model at present. $\star$ indicates tradition methods. - denotes no available result. $\uparrow$ indicates the higher the score the better, and vice versa for $\downarrow$.}
	\label{tab:quantitative}
\end{table*}

\begin{figure*}[t!]
	\centering
	\begin{overpic}[width=0.99\textwidth]{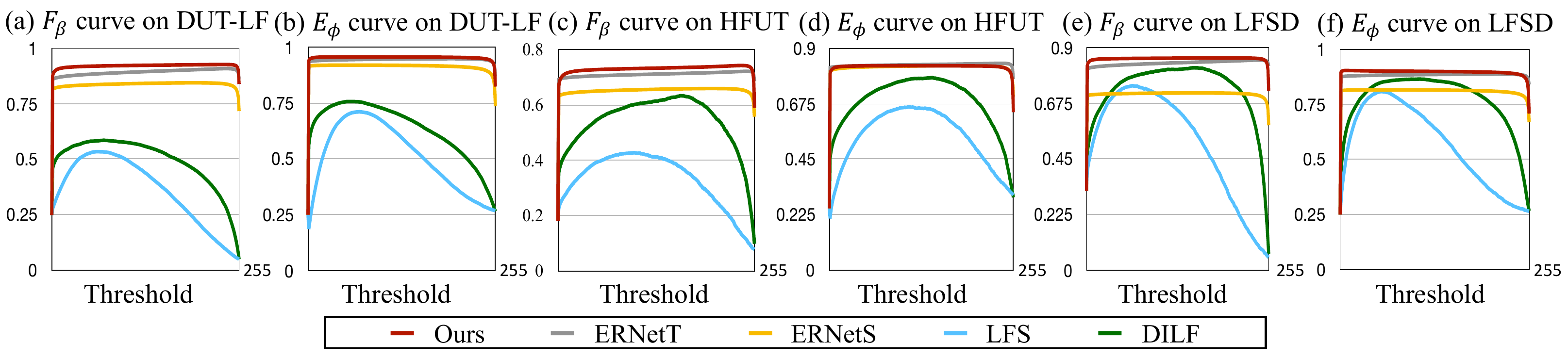}
    \end{overpic}
	\caption{F-Measure ($F_\beta$) and E-Measure ($E_\phi$) curves of cutting edge light field SOD models and our \OurModel~upon three datasets.}
    \label{fig:curves}
\end{figure*}

\subsection{Datasets and Evaluation Metrics}
\noindent\textbf{Datasets}. We evaluate our \OurModel~and \NumOfBaselines~state-of-the-art SOD methods based on three widely-used light field datasets: DUT-LF, HFUT and LFSD, which all provide FS and semantic ground truth corresponding to each of the AiF images (see details in \secref{sec:related_LFSOD}). For fair comparison, we simply follow the settings of a top-ranking method, i.e., ERNet \cite{Piao2020ExploitARERNet}. To be specific, 1000/100 AiF images of DUT-LF/HFUT are randomly selected as the training set, respectively, while the remains (462+155) and the whole LFSD are used for testing. Notably, as for competing methods, we report the results directly provided by authors or generated by officially released codes.

\noindent\textbf{Metrics}. We adopt the newly proposed S-Measure ($S_\alpha$) \cite{fan2017structure} and E-Measure ($E_\phi$) \cite{Fan2018Enhanced}, also the generally agreed Mean Absolute Error ($M$) \cite{perazzi2012saliency} and F-Measure ($F_\beta$) \cite{borji2015salient} as evaluation metrics for the quantitative comparison between benchmark models and \OurModel. Following the benchmark in \cite{Piao2020ExploitARERNet}, we report the adaptive F/E-Measure scores.

\begin{figure*}[t!]
	\centering
	\begin{overpic}[width=0.99\textwidth]{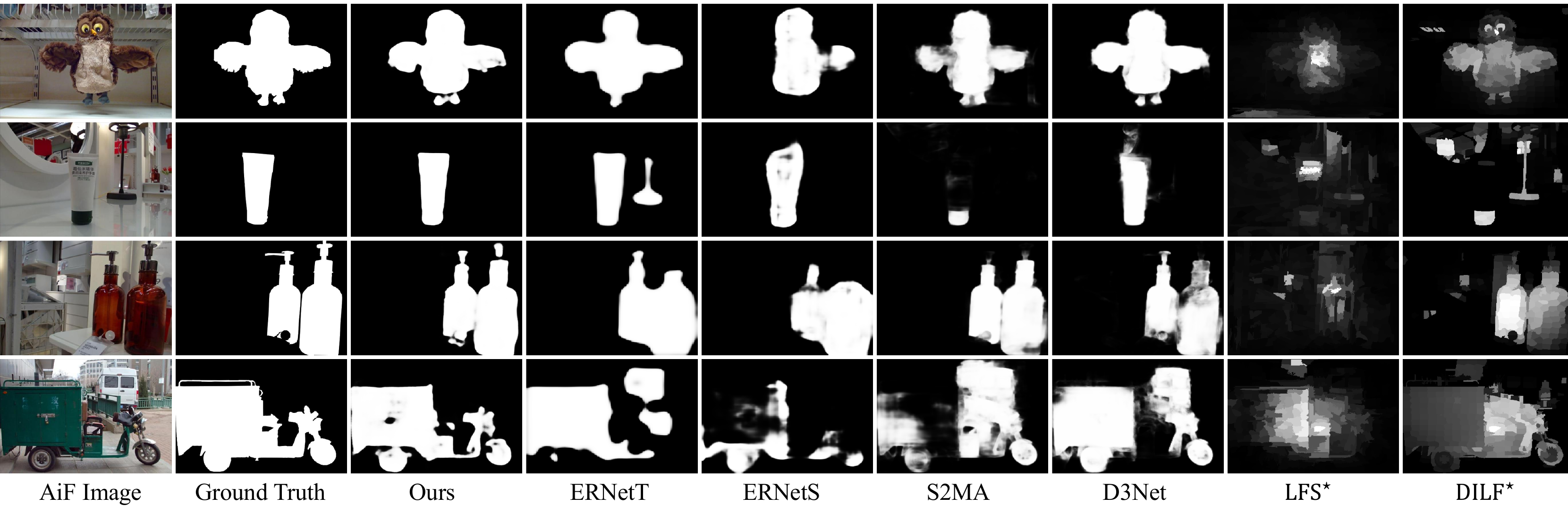}
    \end{overpic}
	\caption{Qualitative comparison between our \OurModel~and state-of-the-art light field SOD models. $\star$ denotes traditional methods. Our \OurModel~provides predictions closest to the ground truth on various aspects.}
    \label{fig:qualitative}
\end{figure*}

\subsection{Comparison with State-of-the-Arts}
We quantitatively compare our \OurModel~with 12/9/7 state-of-the-art RGB/RGB-D/light field SOD methods, respectively.
As shown in Table \ref{tab:quantitative}, our \OurModel~outperforms all the state-of-the-art SOD models by a large margin in terms of all four evaluation metrics.
We also perform a detailed comparison between our \OurModel~and the top-ranking 4D SOD methods by using F/E-Measure curves.
The results, shown in Figure \ref{fig:curves}, indicate the F/E-Measure curves of our \OurModel~are higher than those of competing models.
Furthermore, we show some of the predicted saliency maps in Figure \ref{fig:qualitative}.
As can be observed, our \OurModel~provides the saliency maps closest to the ground truth on various aspects, e.g., correct localization, intact object structure and clear details.
It is worth nothing that our \OurModel~trained on DUT-LF \cite{DUTLF} and HFUT \cite{HFUT} also achieves promising performance on the unseen dataset, i.e., LFSD \cite{LFSD}, indicating its superior generalization ability and robustness. Theoretically, the robustness of \OurModel~owes to the synergistic attention mechanism. In practice, attention mechanisms can improve network robustness \cite{luo2017non,woo2018cbam,zhang2018progressive} since they emphasize the most informative features and reduce the disturbance of noisy features. Our \OurCoAttention~module employs both channel attention and co-attention for better feature representation, which can also improve the robustness of our model.


\subsection{Ablation Studies}\label{sec:ab}
To verify the effectiveness of each proposed module of our \OurModel, we conduct thorough ablation studies by gradually adding key components.
We first construct a baseline ``B'', which extracts AiF and FS features with two 2D ResNet50 backbones, simply concatenates, and up-samples the pair-wise high-level features for SOD.

\noindent\textbf{Effectiveness of multi-modal encoder.}
To investigate the effectiveness of our multi-modal encoder, we construct the second ablated version ``ME'', which is similar to ``B'', but using a 3D backbone to extract FS features, consistent with our multi-model encoder (\secref{sec:encoder}).
The results, shown in Table \ref{tab:ab_quantitative}, indicate that ``ME'' outperforms ``B'' in terms of all evaluations, demonstrating the effectiveness of our 3D CNN-based encoder. Besides, to confirm the effectiveness of RFB for multi-level feature refinement, we also construct ``ME0'' without using the RFB, when compared to ``ME''. The result (Table \ref{tab:ab_quantitative}) shows that RFB benefits significantly to the task.

\begin{wraptable}{l}{0.6\textwidth}
  
	\centering
	\renewcommand{\arraystretch}{0.7}
	\setlength\tabcolsep{2pt}
	\resizebox{0.59\textwidth}{!}{
		\begin{tabular}{cc||ccccccccc}
			\hline
			\toprule
			\multicolumn{2}{c||}{Metric} & B & ME0 & ME & SA1 & SA2 & PF1 & PF2 & F-SA & \OurModel \\
			\midrule
			 \multirow{4}{*}{\begin{sideways}DUT-LF\end{sideways}} &  \multirow{2}{*}{$F_\beta\uparrow$} & \multirow{2}{*}{0.871} & \multirow{2}{*}{0.874} & \multirow{2}{*}{0.881} & \multirow{2}{*}{0.890} & \multirow{2}{*}{0.899} & \multirow{2}{*}{0.912} & \multirow{2}{*}{0.919} & \multirow{2}{*}{0.913} & \multirow{2}{*}{\textbf{0.920}}  \\
			 &&&&&&&&&\\
			 &  \multirow{2}{*}{$M\downarrow$} & \multirow{2}{*}{0.051} & \multirow{2}{*}{0.051} & \multirow{2}{*}{0.048} & \multirow{2}{*}{0.041} & \multirow{2}{*}{0.037} & \multirow{2}{*}{0.035} & \multirow{2}{*}{0.034} &\multirow{2}{*}{0.037} & \multirow{2}{*}{\textbf{0.032}} \\
			 &&&&&&&&&\\
			 	\midrule
			   \multirow{4}{*}{\begin{sideways}LFSD\end{sideways}} &  \multirow{2}{*}{$F_\beta\uparrow$} & \multirow{2}{*}{0.811} & \multirow{2}{*}{0.825}  & \multirow{2}{*}{0.835} & \multirow{2}{*}{0.836} & \multirow{2}{*}{0.835} & \multirow{2}{*}{0.838} & \multirow{2}{*}{0.839} & \multirow{2}{*}{\textbf{0.845}} & \multirow{2}{*}{0.844} \\
			 &&&&&&&&&\\
			 &  \multirow{2}{*}{$M\downarrow$} & \multirow{2}{*}{0.095} & \multirow{2}{*}{0.089} & \multirow{2}{*}{0.080} & \multirow{2}{*}{0.079} & \multirow{2}{*}{0.077} & \multirow{2}{*}{0.075} & \multirow{2}{*}{0.075} & \multirow{2}{*}{0.078} &  \multirow{2}{*}{\textbf{0.074}} \\
			 &&&&&&&&&\\
		
			\bottomrule
		\end{tabular}
	}
	\vspace{5pt}
	\caption{Quantitative results for the ablation studies of \OurModel~on DUT-LF~\protect\cite{DUTLF} and LFSD~\protect\cite{LFSD}. The best scores are in \textbf{boldface}. $\uparrow$ indicates the higher the score the better, and vice versa for $\downarrow$.}
	\label{tab:ab_quantitative}
\end{wraptable}

\begin{wrapfigure}{r}{0.5\textwidth}
\vspace{-7.5mm}
  \begin{center}
    \includegraphics[width=0.49\textwidth]{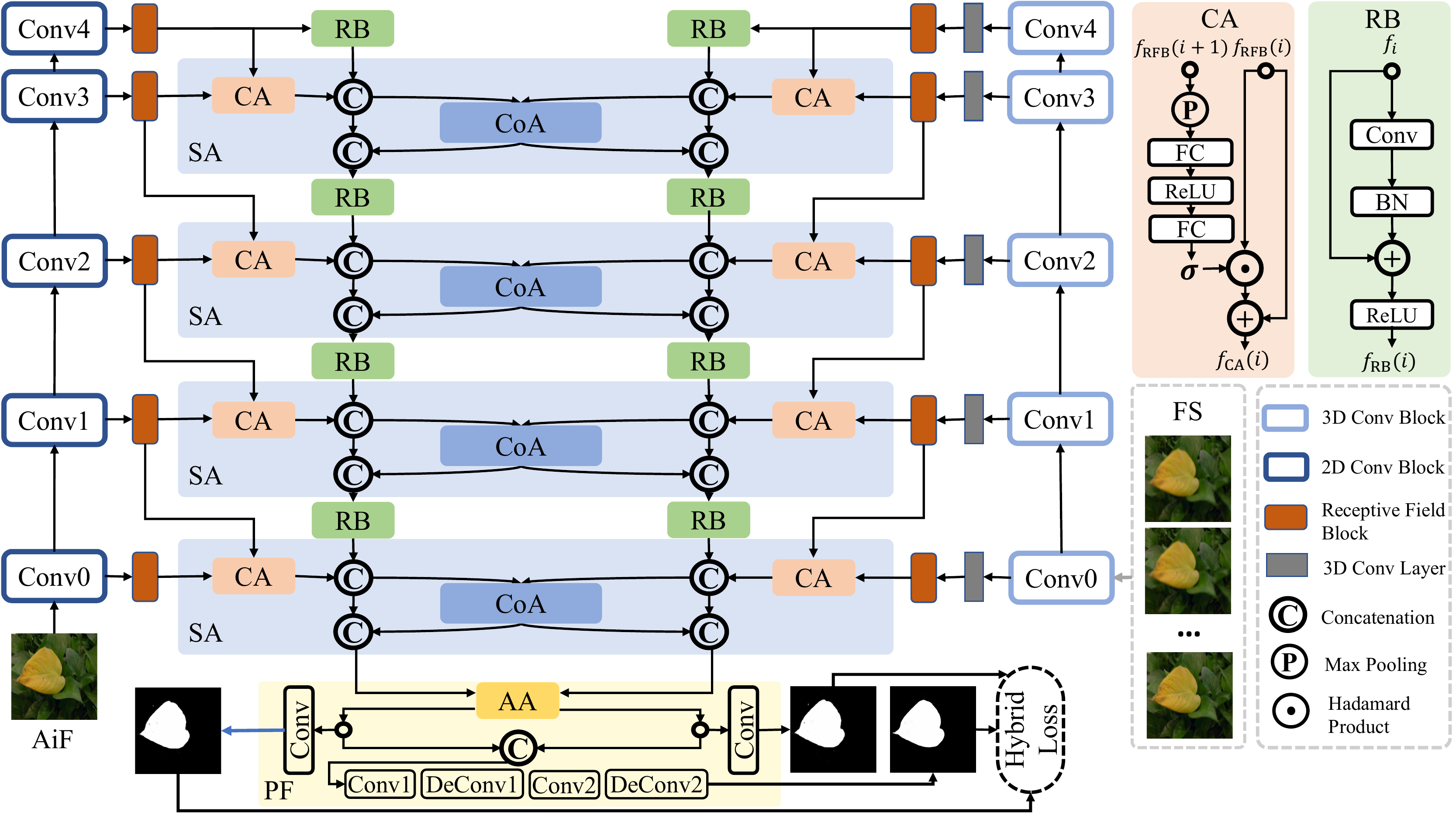}
  \end{center}
  \vspace{-12pt}
  \caption{An overview of F-SA with four \OurCoAttention~modules. CoA = co-attention component. CA = channel attention component. AA = AiF-induced attention component. RB = residual block.}
 \label{fig:model_full_SA}
\end{wrapfigure}

\noindent\textbf{Effectiveness of \OurCoAttention~module.}
To investigate the effectiveness of our \OurCoAttention~module, we further construct ``SA1'' and ``SA2'', which incorporate the \OurCoAttention~into ``ME'' without and with CoA, respectively.
As shown in Table \ref{tab:ab_quantitative}, both ``SA1'' and ``SA2'' improve the performance in comparison with ``ME''.
In particular, the full version of \OurCoAttention~(``SA2'') provides a significant improvement compared to ``ME'', indicating the importance of synergistic attention for learning the complementarities of multi-modal features. 
Besides, we compare \OurModel~with F-SA (Figure \ref{fig:model_full_SA}), which consists of full four \OurCoAttention~modules that fuse both the high and low level features for light field SOD tasks. An interesting finding is that an increase of parameters (about 1.3 million of increment) focusing on low level features do not contribute to performance improvement (Table \ref{tab:ab_quantitative}), which is also consistent with the conclusion in \cite {wu2019cascaded}.

\begin{wrapfigure}{l}{0.5\textwidth}
\vspace{-7.5mm}
  \begin{center}
    \includegraphics[width=0.49\textwidth]{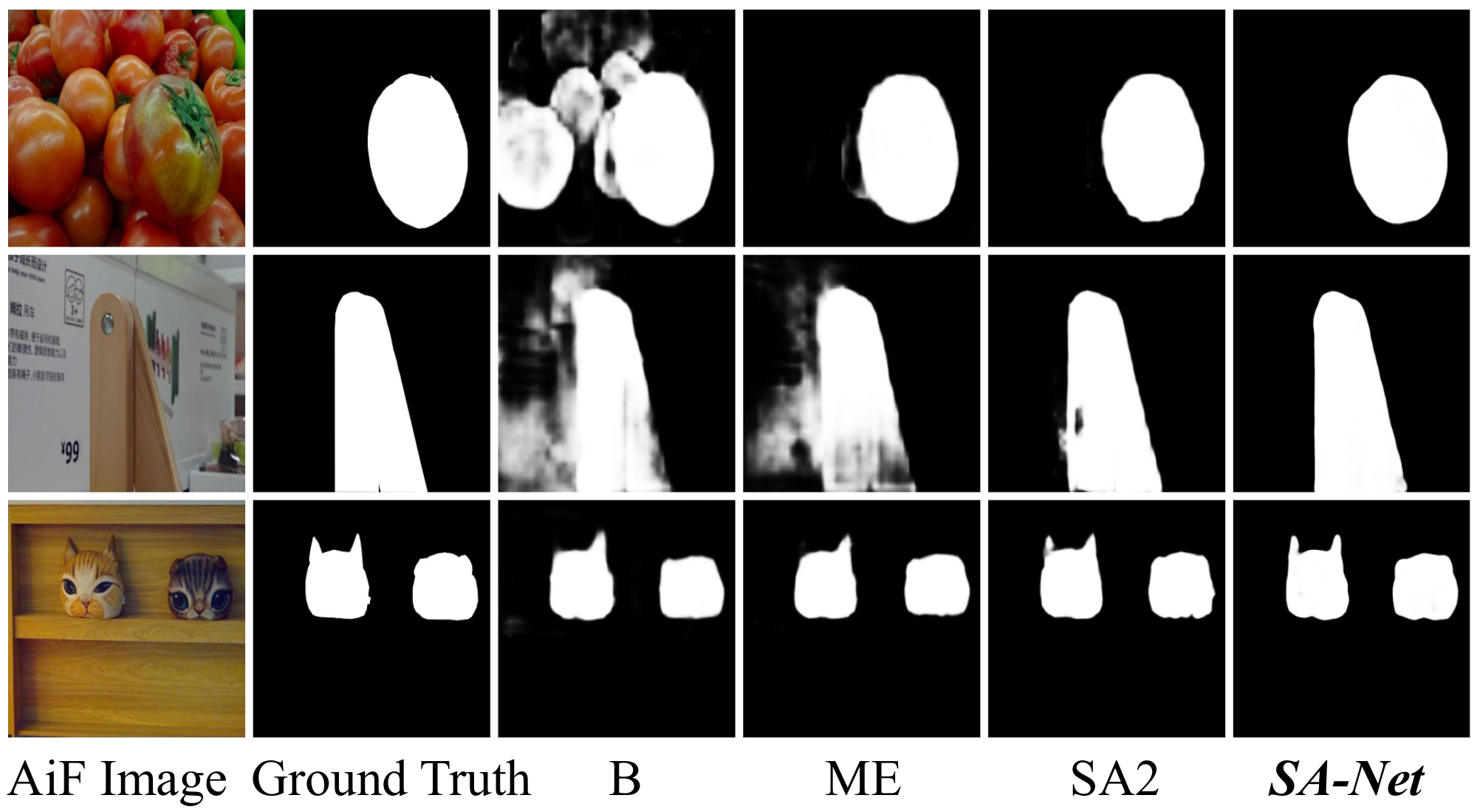}
  \end{center}
  \vspace{-12pt}
  \caption{Representative visual results of ablation studies.
  }
 \label{fig:ab_qualitative}
\end{wrapfigure}

\noindent\textbf{Effectiveness of \OurIntegration~module.}
Compared with ``SA2'', ``PF1'' uses the deconvolutional block (Figure \ref{fig:model_main}) to gradually up-sample the features for predicting the saliency map.
Besides, a three-way supervision (``PF2'') is further employed to provide a deep supervision for the training.
Finally, with the AA component (\secref{sec:PF-module}), our \OurModel~achieves the best performance (Table \ref{tab:ab_quantitative}) and provides the saliency maps closest to ground truth (Figure \ref{fig:ab_qualitative}).

\section{Conclusion}
In this paper, we propose \OurModel, which addresses the light field SOD by learning the synergistic attention for AiF and FS features.
The innovative attributes of our \OurModel~are three-fold:
(i) it exploits the cross-modal complementary information by establishing a synergistic effect between multi-modal features, (ii) it is the first attempt to learn both the spatial and inter-slice features of FSs with 3D CNNs, and (iii) it predicts the saliency map with an effective fusion model in a progressive manner.
Extensive qualitative and quantitative experimental results on three light field datasets demonstrate the better performance of our \OurModel~compared to \NumOfBaselines~competing models. 

\clearpage
\begin{appendices}
\begin{abstract}
This document provides additional information regarding the evaluation metrics and experiments. The four evaluation metrics adopted in the experiments are detailed in this document. Besides, we show more qualitative results to further demonstrate the effectiveness of the proposed \OurModel.
\end{abstract}

\section{Metrics}
In this work, we evaluate all \NumOfBaselines~benchmark models and our \OurModel~with four widely used SOD metrics with respect to the ground-truth binary mask and predicted saliency map. The F-Measure ($F_\beta$) \cite{achanta2009frequency} and mean absolute error (MAE) \cite{perazzi2012saliency} focus on the local (per-pixel) match between ground truth and prediction, while S-Measure ($S_\alpha$) \cite{fan2017structure} pays attention to the object structure similarities. Besides, E-Measure ($E_\phi$) \cite{Fan2018Enhanced} considers both the local and global information.
\par
\noindent
$\bullet$ \textbf{MAE} computes the mean absolute error between the ground truth $G \in \{0, 1\}$ and a normalized predicted saliency map $P \in [0, 1]$, i.e.,
\begin{equation}\label{equ:mae}
   MAE = \frac{1}{W\times{H}}\sum_{i=1}^{W}\sum_{j=1}^{H}\mid G(i, j) - P(i, j)\mid,
\end{equation}
where $H$ and $W$ denotes height and width, respectively.
\par
\noindent
$\bullet$ \textbf{F-Measure} gives a single value ($F_{\beta}$) which considers both the $Precision$ and $Recall$, thus being defined as:
\begin{equation}\label{equ:fm}
   F_{\beta} = \frac{(1+\beta^{2})Precision \times Recall}{\beta^{2}Precision + Recall}, 
\end{equation}
with
\begin{equation}
  Precision=\frac{\left|M\cap G\right|}{\left|M\right|}; Recall=\frac{\left|M\cap G\right|}{\left|G\right|},
\end{equation}
where $M$ denotes a binary mask converted from a predicted saliency map and $G$ is the ground truth. Multiple $M$ are computed by taking different thresholds of $[0,255]$ on the saliency map. Note that the $\beta^{\text{2}}$ is set to 0.3 according to \cite{achanta2009frequency}. Notably, the adaptive F-Measure-based results reported in our manuscript are calculated by applying an adaptive threshold algorithm \cite{borji2015salient}.
\par
\noindent
$\bullet$ \textbf{S-Measure} evaluates the structure similarities between salient objects in ground-truth foreground maps and predicted saliency maps:
\begin{equation}\label{equ:sm}
   S = \alpha \times S_{o} + (1 - \alpha) \times S_{r}.
\end{equation}
where $S_{o}$ and $S_{r}$ denote the object-/region-based structure similarities, respectively. $\alpha \in [0,1]$ is set as 0.5 so that equal weights are assigned to both the object-level and region-level assessments \cite{fan2017structure}.
\par
\noindent
$\bullet$ \textbf{E-Measure} is a cognitive vision-inspired metric to evaluate both the local and global similarities between two binary maps. Specifically, it is defined as:
\begin{equation}\label{equ:em}
E_{\phi}=\frac{1}{W\times H}\sum_{x=1}^W\sum_{y=1}^H\phi\left(P(x,y), G(x,y)\right),
\end{equation} 
where $\phi$ represents the enhanced alignment matrix \cite{Fan2018Enhanced}. Similar to $F_{\beta}$, adaptive E-Measure is adopted for the evaluation in our manuscript.

\section{Qualitative Results}
\begin{figure*}[t!]
	\centering
    \begin{overpic}[width=0.99\textwidth]{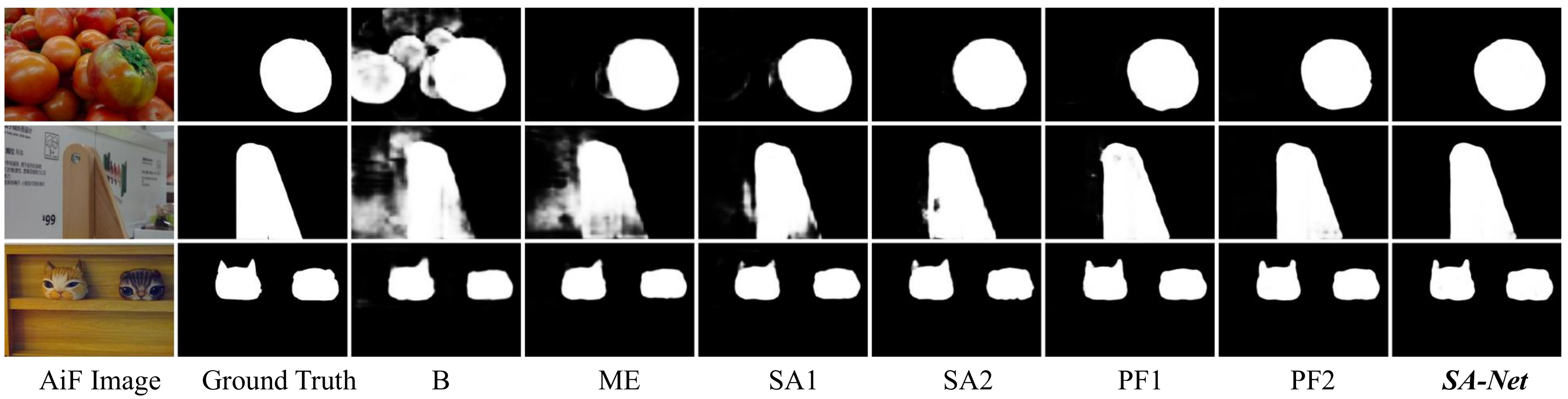}
    \end{overpic}
	\caption{Visual results of ablation studies.}
    \label{fig:supp_ab}
\end{figure*}

\noindent\textbf{Comparison of Ablation Models}. Due to the page limit, we only show partial visual results of ablation studies in our manuscript.
To further illustrate the benefit of each key component in our \OurModel, we show complete qualitative results for all six ablation models in Figure \ref{fig:supp_ab}.
As can be observed, each component improves the quality of predicted saliency maps and contributes to the superior performance of \OurModel.

\begin{figure*}[t!]
	\centering
     {\begin{overpic}[width=0.9\textwidth,trim={0 0 1.8cm 0},clip]{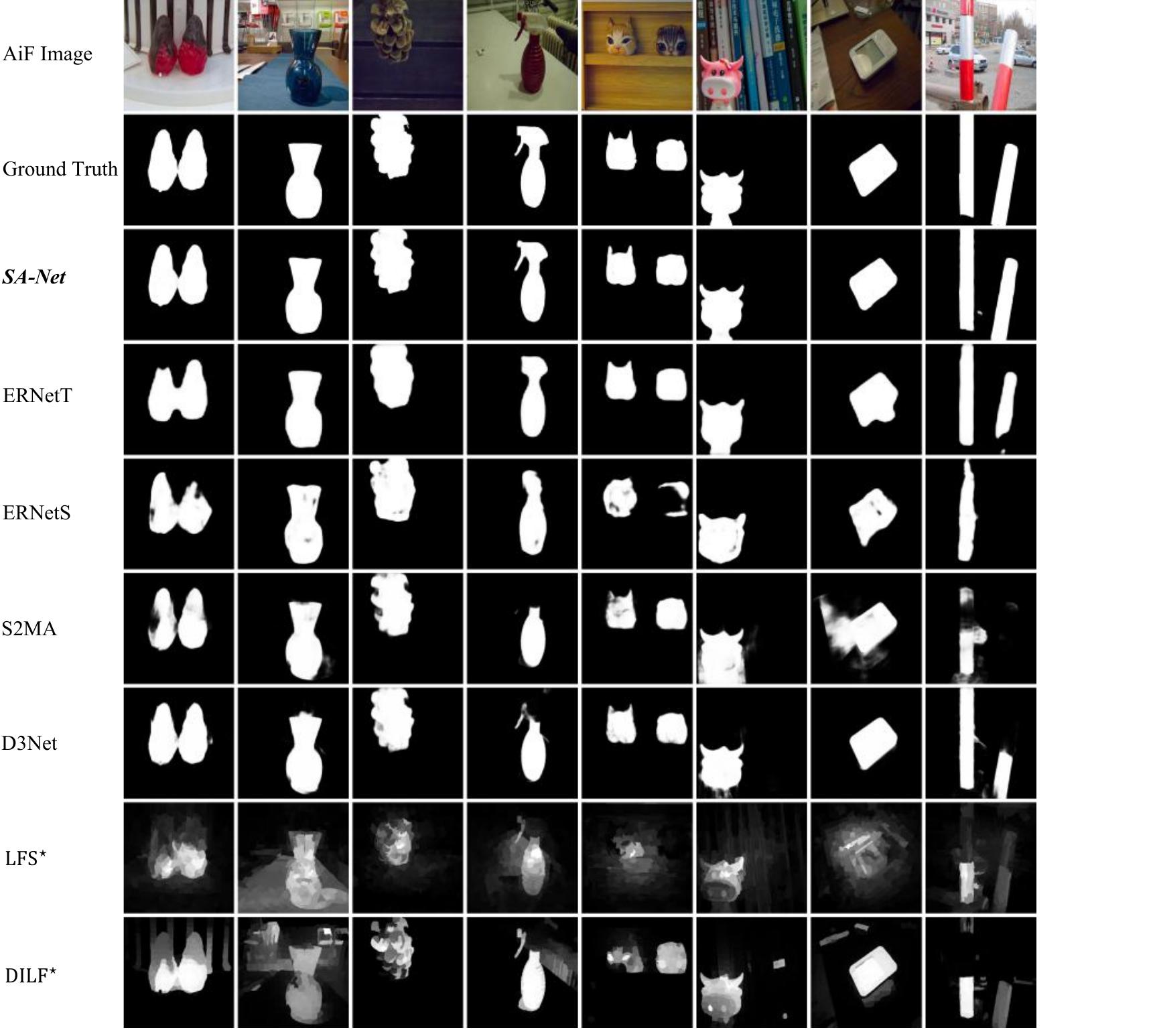} 
    \end{overpic}}
	\caption{Visual comparison of our \OurModel~and state-of-the-art SOD models upon DUT-LF \protect\cite{DUTLF}. $\star$ indicates tradition methods.}
    \label{fig:qualitative_supp_d}
\end{figure*}

\begin{figure*}[t!]
	\centering
     {\begin{overpic}[width=0.9\textwidth,trim={0 0 1.8cm 0},clip]{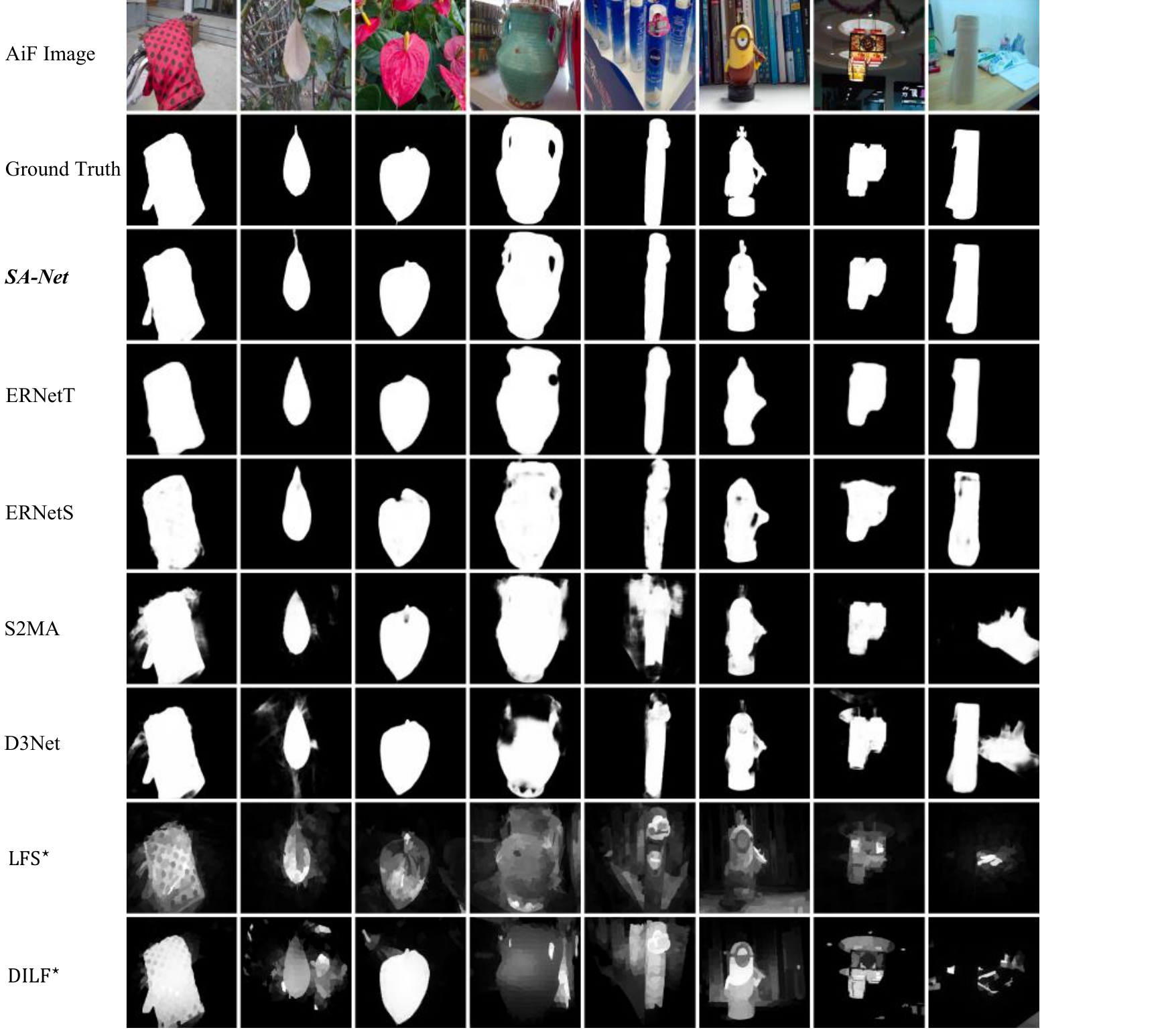} 
    \end{overpic}}
	\caption{Visual comparison of our \OurModel~and state-of-the-art SOD models upon DUT-LF \protect\cite{DUTLF}. $\star$ indicates tradition methods.}
    \label{fig:qualitative_supp_d_2}
\end{figure*}

\begin{figure*}[t!]
	\centering
	\begin{overpic}[width=0.9\textwidth,trim={0 0 1.8cm 0},clip]{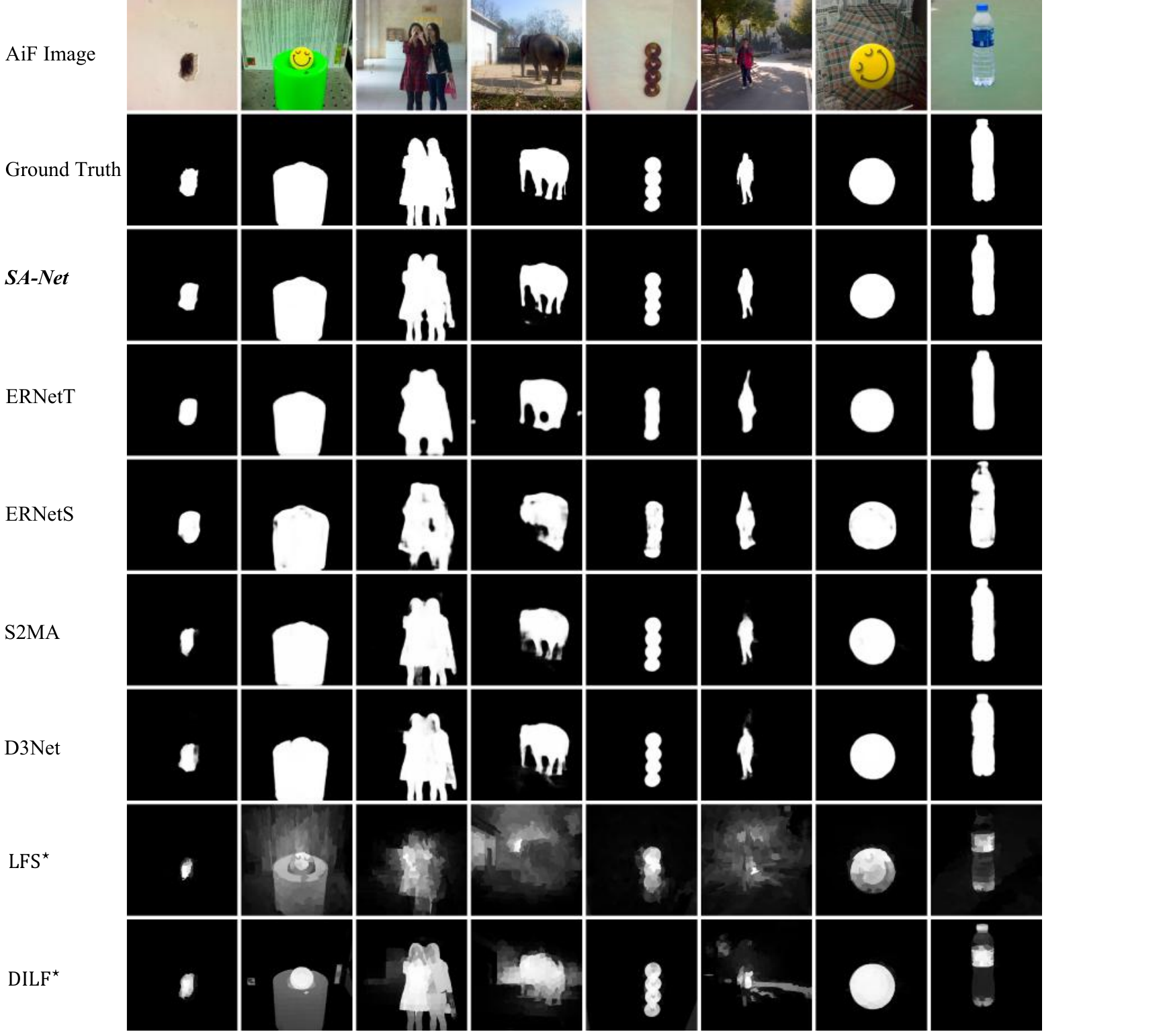}
    \end{overpic}
	\caption{Visual comparison of our \OurModel~and state-of-the-art SOD models upon HFUT \protect\cite{HFUT}. $\star$ indicates tradition methods.}
    \label{fig:qualitative_supp_h}
\end{figure*}

\begin{figure*}[t!]
	\centering
	\begin{overpic}[width=0.9\textwidth,trim={0 0 1.8cm 0},clip]{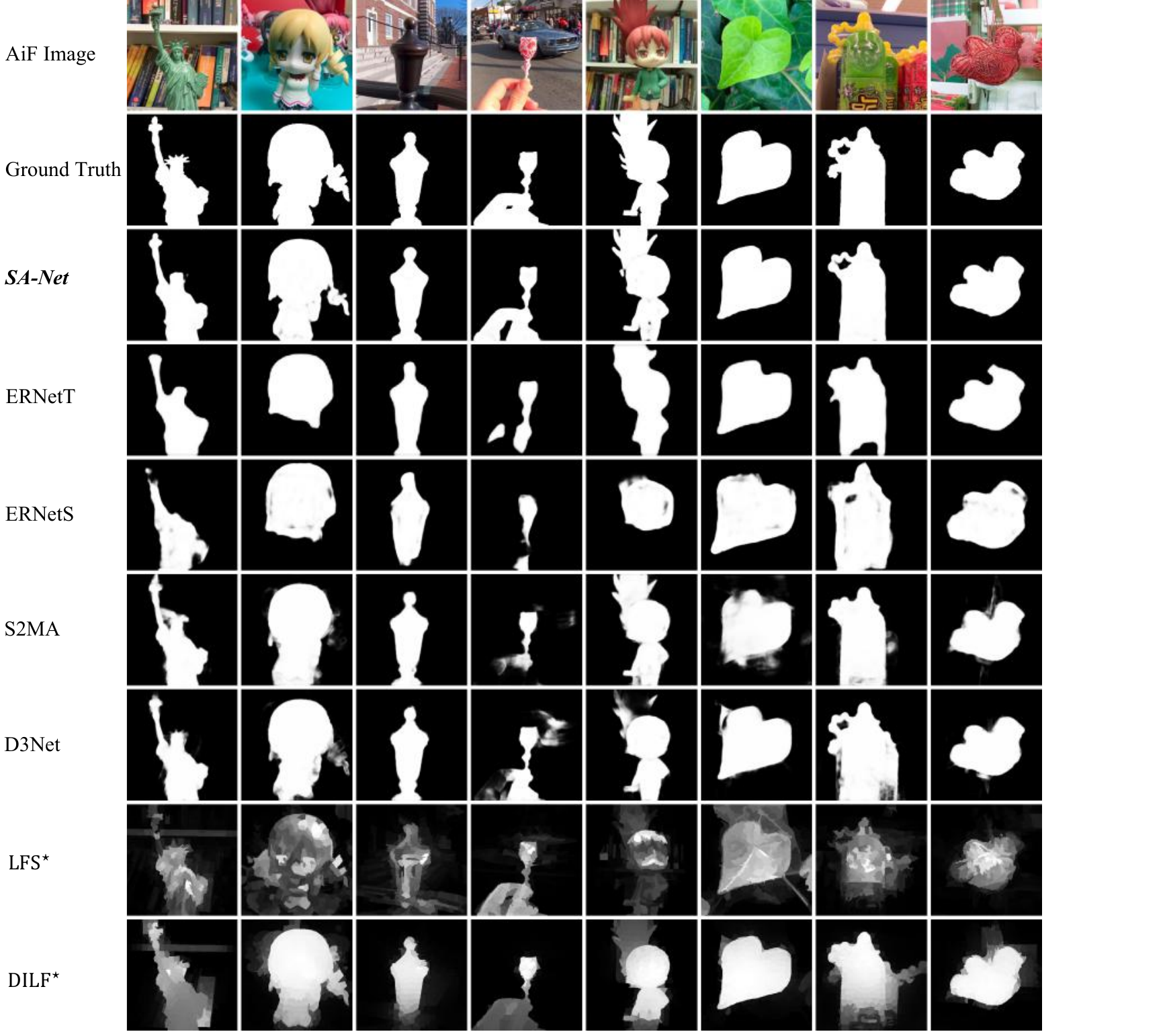}
    \end{overpic}
	\caption{Visual comparison of our \OurModel~and state-of-the-art SOD models upon LFSD \protect\cite{LFSD}. $\star$ indicates tradition methods.}
    \label{fig:qualitative_supp_l}
\end{figure*}

\noindent\textbf{Comparison with State-of-the-Arts}. To further demonstrate the effectiveness of our \OurModel, we show extensive visual results of our method as well as the competing models upon the three benchmark datasets (Figure \ref{fig:qualitative_supp_d} to \ref{fig:qualitative_supp_l}). Overall, our proposed \OurModel~depicts fine object structures and possesses less false positive and false negative, thus giving predictions closest to ground truths.

\clearpage

\clearpage

\clearpage

\clearpage
\end{appendices}

\clearpage
\bibliography{egbib}
\end{document}